\documentclass{article} 
\usepackage{iclr2015,times}
\usepackage{hyperref}
\usepackage{url}
\usepackage{times}
\usepackage{epsfig}
\usepackage{graphicx}
\usepackage{subfigure}
\usepackage{amsmath}
\usepackage{tabularx}
\usepackage{amssymb}
\usepackage{mathrsfs}
\usepackage{algorithm}
\usepackage{algorithmic}
\usepackage{multirow}
\usepackage{multicol}
\usepackage{bm}

\title{Real-World Font Recognition Using Deep Network and Domain Adaptation}

\author{
Zhangyang Wang \& Thomas S. Huang\\
Beckman Institute for Advanced Science and Technology \\
University of Illinois at Urbana-Champaign\\
Urbana, IL 61801, USA \\
\texttt{\{zwang119, t-huang1\}@illinois.edu} \\
\And
Jianchao Yang \& Hailin Jin \& Eli Shechtman \&  Aseem Agarwala \& Jonathan Brandt\\
Adobe Research \\
San Jose, CA, 95110\\
\texttt{\{jiayang, hljin, elishe, asagarwa, jbrandt\}@adobe.com} \\
}

\iclrfinalcopy 

\begin{document}

\maketitle

\begin{abstract}
We address a challenging fine-grain classification problem: recognizing a font style from an image of text. In this task, it is very easy to generate lots of rendered font examples but very hard to obtain real-world labeled images. This real-to-synthetic domain gap caused poor generalization to new real data in previous methods (\cite{LFE}). In this paper, we refer to Convolutional Neural Networks, and use an adaptation technique based on a Stacked Convolutional Auto-Encoder that exploits unlabeled real-world images combined with synthetic data. The proposed method achieves an accuracy of higher than 80\% (top-5) on a real-world dataset.
\end{abstract}

\section{Introduction}

This paper studies font recognition, i.e. identifying a particular typeface given an image of a text fragment. To apply machine learning to this problem, we require realistic text images with ground truth font labels.  However, such data is scarce and expensive to obtain, since it requires a high level of domain expertise which is out of reach of most people. Therefore, it is infeasible to collect a sufficient set of real-world training images. One way to overcome the training data challenge is to synthesize the training set by rendering text fragments for all the necessary fonts. However, we must face the domain mismatch between synthetic and real-world text images (\cite{LFE}). Characters in real-world images are spaced, stretched and distorted in numerous ways.  In (\cite{LFE}), the authors tried to overcome this difficulty by adding different degradations to synthetic data.  In the end, introducing all possible real-world degradations into the training data is infeasible. 



We address this domain mismatch problem in font recognition, by further leveraging a large corpus of synthetic data to train a Convolutional Neural Network (CNN), while introducing an adaptation technique based on Stacked Convolutional Auto-Encoder (SCAE) with the help of unlabeled real-world images. The proposed method reaches an impressive performance on real-world test images.

\section{Model}



\begin{figure*}[htbp]
\centering
\begin{minipage}{0.90\textwidth}
\centering {
\includegraphics[width=\textwidth]{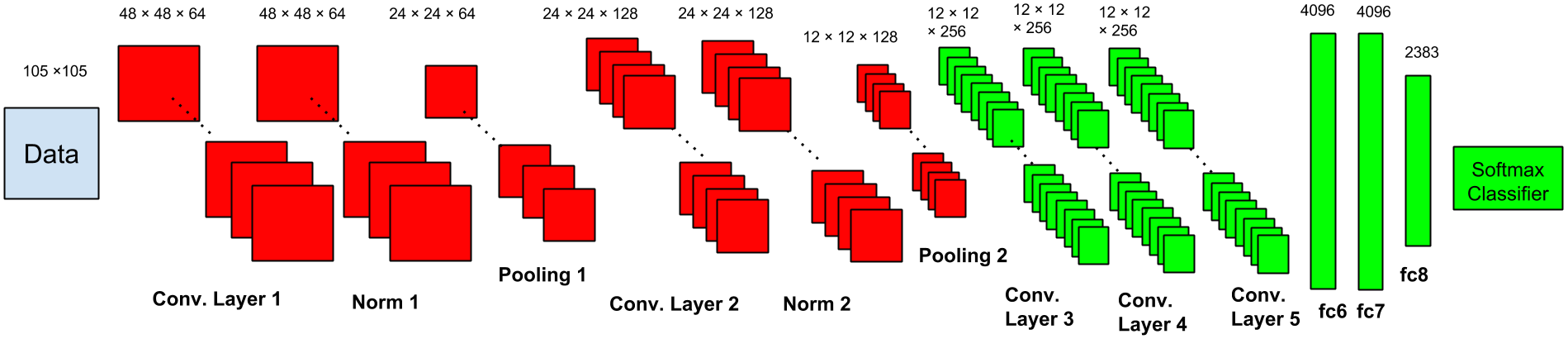}
}\end{minipage}
\caption{The CNN architecture, and its decomposition marked by different colors ($N$=8, $K$=2).}
\label{fig:CNN}
\end{figure*}

Our basic CNN architecture is similar to the popular ImageNet CNN structure in (\cite{imagenet}), as depicted in Fig. \ref{fig:CNN}. The numbers along with the network pipeline specify the dimensions of outputs of corresponding layers.  When the CNN model trained fully on a synthetic dataset, it witnesses a significant performance drop when testing on real-world data, compared to when applied to another synthetic validation set. This also happens with other models such as in \cite{LFE}, which uses training and testing sets of similar properties to ours. This alludes to discrepancies between the distributions of synthetic and and real-world examples.


Tradition approaches to handle this gap include pre-processing steps applied on the training and/or testing data (\cite{LFE}). The domain adaptation method in \cite{sentiment} extracts low-level features that represent both the synthetic and real-world data, based on a stacked auto-encoder (SAE). We extend the method in \cite{sentiment} to decompose the $N$ basic CNN layers into two sub-network parts. The first $K$ layers accounts for extracting low-level visual features shared by both synthetic and real-world data, and will be learned in a unsupervised way, using unlabeled data from both domains. The remaining $N-K$ layers accounts for learning higher-level discriminative features for classification. It will be trained in a supervised way on top of the first part, using labeled data from the synthetic domain only.

To train the first $K$ layers, we exploit a Stacked Convolutional Auto-Encoder (SCAE) (\cite{SCAE}). Its first two convolutional layers have an identical topology to the first two layers in Fig. \ref{fig:CNN}. Moreover, we set its first and second half to be mirror-symmetrical. The cost function is the mean squared error (MSE) between the input and reconstructed patches. After SCAE is learned, its Conv. Layers 1 and 2 are imported to the CNN in Fig. \ref{fig:CNN}. We adopt the SCAE implementation by \cite{paine2014analysis}.


We also find that applying label-preserving data augmentations to synthetic training data helps reduce the domain mismatch. \cite{LFE} added moderate distortions and corruptions, including noise, blur, rotations and shading effects. In addition, we also vary the character spacings and aspect ratios when rendering training data. Note that these steps are not useful for the method in \cite{LFE} because it exploits very localized features, but they are very helpful in our case.

\section{Experiments}

We implemented and evaluated the local feature embedding-based algorithm (LFE) in (\cite{LFE}) as a baseline, and compare it with our model. A SCAE is first trained on a large collection of both synthetic data and unlabeled real world data, and then exports the first $K = 2$ convolutional layers. The next $N - K$ layers are trained on labeled synthetic data covering 2,383 classes. That makes our problem quite fine-grain. Testing is conducted on the the VFRWild325 dataset used by (\cite{LFE}), in term of top-1 and top-5 classification errors. Our model achieves 38.15\% in top-1 error and 20.62\% in top-5, which outperforms 6\% and 10\% over LFE, respectively.

\bibliography{iclr2015}
\bibliographystyle{iclr2015}

\end{document}